\documentclass[10pt,twocolumn,letterpaper]{article}

\usepackage{cvpr}              

\usepackage{notoccite}

\usepackage{graphicx}
\usepackage{amsmath}
\usepackage{amssymb}
\usepackage{booktabs}
\usepackage[T1]{fontenc}
\usepackage{multirow}

\usepackage[pagebackref,breaklinks,colorlinks]{hyperref}
\usepackage{amsmath}

\usepackage[capitalize]{cleveref}
\crefname{section}{Sec.}{Secs.}
\Crefname{section}{Section}{Sections}
\Crefname{table}{Table}{Tables}
\crefname{table}{Tab.}{Tabs.}

\def\cvprPaperID{52} 
\def\confName{CVPR}
\def\confYear{2022}

\begin{document}

\title{SqueezeNeRF: Further factorized FastNeRF for memory-efficient inference}

\author{Krishna Wadhwani\\
Sony Group Corporation\\
{\tt\small krishna.a.wadhwani@sony.com}
\and
Tamaki Kojima\\
Sony Group Corporation\\
{\tt\small tamaki.kojima@sony.com}
}
\maketitle

\begin{abstract}
Neural Radiance Fields (NeRF) has emerged as the state-of-the-art method for novel view generation of complex scenes, but is very slow during inference. Recently, there have been multiple works on speeding up NeRF inference, but the state of the art methods for real-time NeRF inference rely on caching the neural network output, which occupies several giga-bytes of disk space that limits their real-world applicability. As caching the neural network of original NeRF network is not feasible, Garbin et.al. proposed "FastNeRF" which factorizes the problem into 2 sub-networks - one which depends only on the 3D coordinate of a sample point and one which depends only on the 2D camera viewing direction. Although this factorization enables them to reduce the cache size and perform inference at over 200 frames per second, the memory overhead is still substantial. In this work, we propose SqueezeNeRF, which is more than 60 times memory-efficient than the sparse cache of FastNeRF and  is still able to render at more than 190 frames per second on a high spec GPU during inference.
\end{abstract}
\section{Introduction}
\label{sec:intro}
Image-based rendering and novel view synthesis are fundamental problems in computer vision, and there is a rich and long history of research works in these directions \cite{old_ref_1, quicktimevr, old_ref_2}. This field of novel view synthesis, using a set of observed images of a scene to recover 3D representation of the scene that allows rendering from unobserved viewpoints, has seen an unprecedented rise in popularity with the proposal of Neural Radiance Fields (NeRF) \cite{nerf}. Given a limited amount of images of a scene, NeRF learns an implicit volumetric representation of the scene that allows photo-realistic rendering of novel views of the scene that is able to capture fine level details and view-dependent effects. Essentially, it is a multi-layer perceptron (MLP) based model to learn a mapping of 5D input-3D coordinates plus 2D viewing directions to density and color values. The model is optimized on a set of training views of a scene with known camera poses. The learnt model can then be used to render novel views by volume rendering.

However, despite impressive view generation results, NeRF has several limitations. Some of them are: (i)~ it cannot handle dynamic objects and scene appearance(s), (ii)~it cannot handle unbounded scenes, (iii)~a network trained on one scene cannot generalize across another scene, (iv)~it requires relatively large number of views to learn an accurate representation, and  (v)~slow inference speed. These several limitations of NeRF have inspired several follow-up works such as~\cite{nerfpp, dsnerf,pixelnerf, plenoxels}. 

\begin{figure}[t!]
  \centering
  \includegraphics[width=0.95\linewidth]{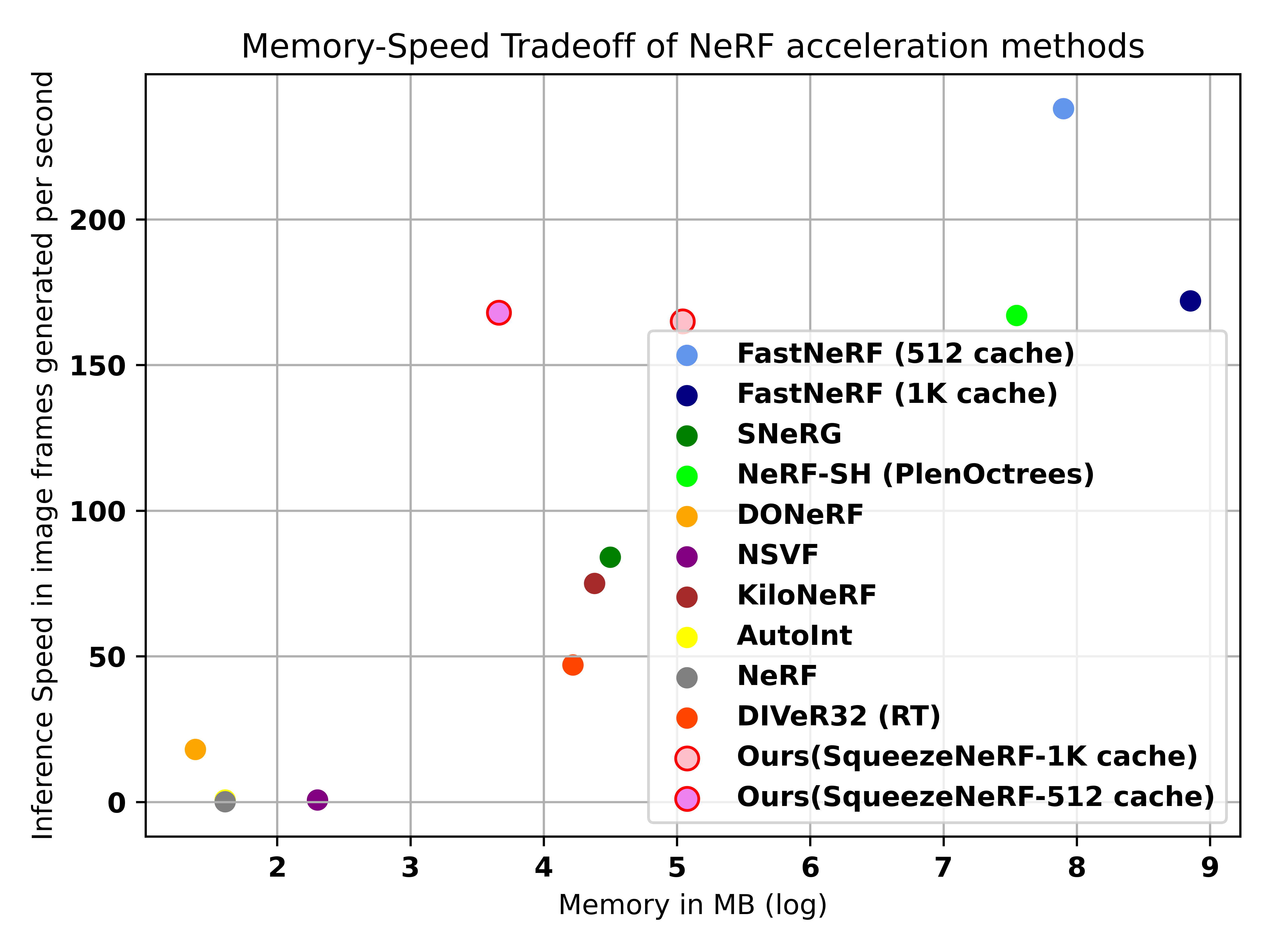}
   \caption{Memory requirement vs inference speed (to generate 800x800 image) for faster inference based NeRF models on the synthetic 360$^\circ$ dataset\cite{nerf}. Note: Reported numbers in this figure may have different GPU settings. FastNeRF result uses Nvidia RTX-3090 while PlenOctrees and SqueezeNeRF have been tested on Nvidia V100.}
   \label{fig:mem_speed_graph}
\end{figure}

In this work, we focus on the slow rendering issue of NeRF. The reason why NeRF is slow during inference is also, its greatest strength - an MLP-based volumetric representation. How NeRF works is that for a given camera pose, a ray is projected that passes through each pixel in the image into the 3D scene. We then sample points on each ray, predict the color and density value at each point and then, do volumetric integration along each ray to compute the final color at the pixel location. While the simplicity of NeRF's representation of a 3D scene via a MLP is elegant and results in high-quality novel view synthesis of a scene, it also means that during inference, we need to query the neural network to get the color and density values for millions of points. For example, in order to render a 800x800 image by sampling 192 points along each ray, we will need to query the neural network for 122.88 million 3D points. This means that it takes a several seconds to render an image via NeRF on a high-end GPU.  

There have been multiple works that have attempted to speed-up NeRF inference \cite{autoint, nsvf, fastnerf, snerg, kilonerf, coordx, plenoctrees, diver}. While all of them are considerably faster than the original NeRF model during inference, they either still do not achieve real time rendering \cite{autoint, nsvf, coordx} or achieve real-time rendering at the cost of significant memory overhead \cite{fastnerf, snerg, plenoctrees, diver}.

The state-of-the-art approaches in speeding up NeRF inference rely on caching the mapping learnt by MLP. Now it must be stated that caching the original NeRF model is not practically possible. This is because NeRF maps a 5-dimensional input (3D coordinate $\boldsymbol{p}$ $+$ 2D viewing direction $\boldsymbol{d}$) to a 4-dimensional output (RGB color $\boldsymbol{c}$ and scalar density $\sigma$). So caching this function in the input space would mean that the cache-size would scale with $n^5$ (where $n$ is the number of bins or sampled values per input dimension). For $n=512$ bins per input dimension, the cache size would be approximately 176 terabytes and infeasible to store. In order to solve this issue, FastNeRF\cite{fastnerf} proposed a novel architecture that factorizes the problem into 2 independent networks - their first MLP takes the 3D coordinate $\boldsymbol{p}$ as input and the second MLP takes the 2D viewing direction $\boldsymbol{d}$ as input. The outputs of these two networks are combined via their inner product to produce the RGB color values. Now considering $n_p$ number of bins for each position dimension and $n_d$ number of bins for each viewing direction component, the FastNeRF cache size would have a complexity of $\mathcal{O}(D_pn_p^3 + D_dn_d^2) \sim \mathcal{O}(n^3)$ where $D_p$ and $D_d$ represent the dimensionality of the position dependent MLP output and viewing direction dependent MLP output respectively. For standard FastNeRF configuration of $D_p=25, D_d=8$ and $n_p=512, n_d=256$, the cache size is approximately 6.7 GB, if all the values are stored as half-precision floating point numbers. The FastNeRF authors further leveraged the scene sparsity to store a sparse cache instead of a dense one in order to reduce the cache size to around 3GB. Similarly, NeRF-SH~\cite{plenoctrees} and SNeRG~\cite{snerg} also use an architecture based on similar factorization but rely on different architecture and cache-storage strategy. 

While FastNeRF and NeRF-SH (PlenOctrees based inference) can render over 150 frames per second (FPS) on a high-end GPU, their cache takes up several gigabytes (GBs) of memory, which make them prohibitive for any embedded system application and can lead to memory fragmentation issues during inference. Moreover, as a NeRF requires a different MLP to be trained for each scene, storing several GBs of cache for every scene makes this memory overhead issue even worse. 

To address this issue, we propose SqueezeNeRF, a further factorized version of FastNeRF which is able to render over 150 FPS with affordable cache-size. This makes the model convenient for deploying on embedded systems. SqueezeNeRF factorizes the position dependent MLP of FastNeRF into 3 separate MLPs - which take $(x,y), (y,z)$ and $(z,x)$ as input. This factorization enables SqueezeNeRF to have a cache-size complexity of $\mathcal{O}((D_{p_{xy}} +D_{p_{yz}} + D_{p_{zx}})n_p^2 + D_dn_d^2) \sim \mathcal{O}(n^2)$, where $D_{p_{xy}}, D_{p_{yz}}, D_{p_{zx}}$ represent the dimensionality of the 3 position dependent networks. For a standard SqueezeNeRF configuration of $D_{p_{yz}}=D_{p_{yz}}=D_{p_{yz}}=25, D_d=8$ and $n_p=512, n_d=64$, the cache-size is approximately 40 MB, which is more than 160 times memory efficient than the dense FastNeRF cache, more than 65 times memory efficient that sparse FastNeRF cache. We further compare our approach with the fast inference based NeRF approaches in terms of performance, speed and memory requirements and show that our relatively straightforward approach is competitive with the state-of-the-art approaches in terms of speed while being more memory-efficient, although is slightly inferior in terms of image generation performance. 

The contributions of our work can be summarized as:
\begin{itemize}
    \item We propose SqueezeNeRF, which is the first approach that allows NeRF rendering at over 150 frames per second with cache-size less than 200 megabytes. The approach is based on a novel NeRF architecture, that is based on further factorization of FastNeRF.
    \item We present a comprehensive evaluation of NeRF follow-up works for faster inference and compare their speed, performance and memory requirements. 
\end{itemize}

\section{Related Work}
\label{sec:related}

Neural Radiance Fields~\cite{nerf} has shown impressive results in the task of novel view generation of a static scene but is limited by its slow inference speed. A number of recent works~\cite{autoint, nsvf, fastnerf, snerg, kilonerf, coordx, plenoctrees, nex} have tried to address this particular limitation of NeRF. Our proposed model, SqueezeNeRF also belongs to these family of methods aimed to improve the inference speed of NeRF. 


AutoInt \cite{autoint} proposed a neural network to learn integrals along each ray with reduces number of point samples on each ray.
Neural Sparse Voxel Fields~\cite{nsvf} learns a sparse voxel grid that allows to skip over empty region during inference.
%
%
While considerably faster than NeRF,~\cite{autoint, nsvf} do not render at interactive frame rates.

KiloNeRF~\cite{kilonerf} splits one network into multiple tiny networks and parallelizes the multiple network evaluations. 
DIVeR~\cite{diver} combines implicit representation learning with voxel based representation and uses deterministic integration instead of the sampling scheme done in NeRF. 
DONeRF \cite{donerf} predicts depth distribution along each ray for using reduced number of samples in volumetric rendering. 
NeX~\cite{nex} models the scene as a multi-plane image (MPI) rather than a continuous volumetric representation. While NeX shows impressive results, it cannot model scenes where the viewpoint span is high. 
State-of-the-art methods \cite{fastnerf, plenoctrees, snerg} in fast inference of NeRF rely on caching the neural network output. Caching NeRF MLP is not feasible due to extremely high memory complexity. So all these work modify the NeRF architecture in order to make this caching possible. 
All ~\cite{plenoctrees, fastnerf, snerg} factorize the structure of the implicit model, by using two separate neural networks for 3D position and camera view direction. While this factorization allows their network cache to be 1000 times smaller than NeRF and perform inference at interactive rate, their cache-size is still in the order of few giga-bytes. Although memory requirement in \cite{kilonerf, diver, donerf, snerg} is smaller due to not relying on cache \cite{kilonerf, donerf} or efficient compression scheme \cite{snerg}, their rendering speed is less than those in FastNeRF \cite{fastnerf} and NeRF-SH \cite{plenoctrees}. In this work, we further factorize the position dependent network in FastNeRF which allows us to reduce the FastNeRF cache-size by more than 60 times as compared to FastNeRF or NeRF-SH, while still maintaining similar rendering speed. 
%

A concurrent work \cite{coordx} also proposed to combine implicit representation based learning neural networks with a split MLP architecture, which is similar in spirit to the idea of factorization used by FastNeRF or us. They use a separate MLP for each dimension of the input but they do not employ caching for faster inference and their speed-up for inference is significantly less than our approach.

\section{Method}
\label{sec:method}

\begin{figure*}[!t]
    \centering
    \includegraphics[width=0.8\linewidth]{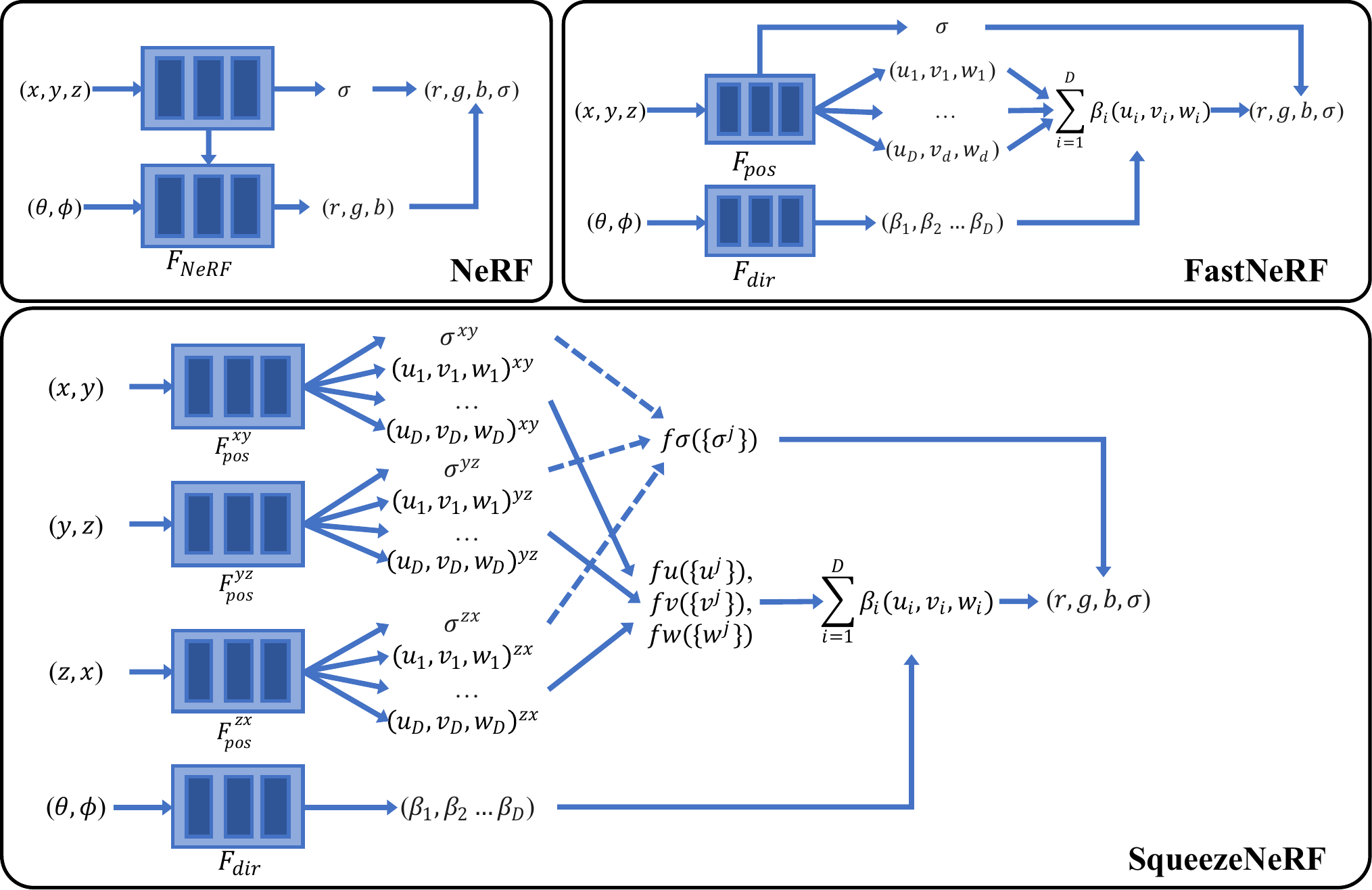}
    \caption{\textbf{Neural network architecture comparison.} Top left: NeRF architecture. Cache-size of the original NeRF network is of the order $\mathcal{O}(n^5)$, where $n$ is the number of bins per input dimension. Top right: FastNeRF \cite{fastnerf} architecture based on factorization into position dependent network $F_{\text{pos}}$ and $F_{\text{dir}}$. Cache-size of FastNeRF network is of the order $\mathcal{O}(n^3)$. Bottom: Proposed SqueezeNeRF architecture that further factorizes the position dependent network in FastNeRF into 3 networks: $F_{\text{pos}}^{xy}$, $F_{\text{pos}}^{yz}$ and $F_{\text{pos}}^{zx}$. Cache-size of SqueezeNeRF network is of the order $\mathcal{O}(n^2)$. In the above diagram, $(x,y,z)$ denotes 3D position of input sample, $(\theta,\phi)$ denotes camera ray direction and $(r,g,b,\sigma)$ are the color and density/opacity values.}
    \label{fig:arc}
\end{figure*}

In this section, we describe SqueezeNeRF. FastNeRF splits NeRF into 2 neural networks - a position $(x,y,z)$ dependent network and viewing direction dependent network. The key idea in this work is to further split the position dependent network into 3 networks, which are dependent on $(x,y)$, $(y,z)$ and $(z,x)$ respectively. We briefly recap NeRF and FastNeRF in the following subsection before describing the details of our model.

\subsection{Background}
\label{sec:nerf}

\textbf{Neural Radiance Fields (NeRF)}: Given multi-view images of a scene with known camera poses and parameters, NeRF \cite{nerf} learns to recover the 3D volumetric representation of a scene in the form of (i)~opacity field or volume density $\sigma$ that captures a soft or an approximate geometry of the scene and (ii)~a radiance field or RGB color $\boldsymbol{c}$ that captures the view dependent surface texture. This representation is captured in the form of a mapping from 3D coordinate position $\boldsymbol{p} = (x,y,z)$ and 2D viewing direction $\boldsymbol{d} = (\theta, \phi)$. A MLP $F_{NeRF}$ is used to learn this mapping $F_{NeRF}: (\boldsymbol{p}, \boldsymbol{d}) \rightarrow (\sigma, \boldsymbol{c})$. The density $\sigma$ is modelled as a function of $\boldsymbol{p}$. On the other hand, the color $\boldsymbol{c}$ is modelled as a function of both $\boldsymbol{p}$ and $\boldsymbol{d}$. 

How this process works is that for a given viewing angle, in order to render a single pixel at pixel location $\boldsymbol{P}$, a ray is projected from the camera center that passes through that pixel and into the 3D scene. This ray direction is denoted as $\boldsymbol{d}$. $N$ number of 3D points are sampled along this ray ($\boldsymbol{p}_1, \boldsymbol{p}_2 \dots \boldsymbol{p}_N $) between the near and far plane of the camera. Each point sample is fed as input to the neural network which then predicts the color $\boldsymbol{c}_i$ and density $\sigma_i$ at each point $\boldsymbol{p}_i$. The color and density value at each point is then used to compute the final pixel color $\hat{\boldsymbol{c}}$ using volumetric integration as: 
\begin{equation}
    \hat{\boldsymbol{c}}(\boldsymbol{P}) = \sum_{i=1}^N T_i(1-\text{exp}(-\sigma_i\delta_i )c_i),
     \label{eq:vol}
\end{equation}
where $T_i=\text{exp}(-\sum_{j=1}^{i-1}\sigma_j\delta_j)$ is the transmittance and $\delta_i=(\boldsymbol{p}_{i+1}-\boldsymbol{p}_i)$ is the distance between adjacent samples.

Given a set of training images with known camera poses, NeRF optimizes the MLP by minimizing the squared error between input pixel color and the pixel color value predicted using \cref{eq:vol}. So, the network weights are trained to optimize:

$L_p = \sum_{\boldsymbol{P}} \vert\vert \boldsymbol{c}(\boldsymbol{P}) - \hat{\boldsymbol{c}}(\boldsymbol{P}) \vert\vert_2^2,$

where $\boldsymbol{c}(\boldsymbol{P})$ is the ground truth color at $\boldsymbol{P}$. Hence, by replacing an explicit volumetric representation with an MLP, NeRF requires orders of magnitude less space than a dense voxel grid, but rendering an image requires querying the neural network at millions of 3D points, which makes the rendering process slow and computationally expensive.

\textbf{FastNeRF}: As mentioned earlier, FastNeRF\cite{fastnerf} splits NeRF's neural network $F_{NeRF}$ into two networks, (i)~position dependent network $F_{\text{pos} }:~\boldsymbol{p}~\rightarrow~ \{ \sigma, (\boldsymbol{u}, \boldsymbol{v}, \boldsymbol{w})\}$ and (ii)~ray direction dependent network $F_{\text{dir}}:~\boldsymbol{d} \rightarrow \boldsymbol{\beta}$ where $\boldsymbol{u}, \boldsymbol{v}, \boldsymbol{w}$ are $D$-dimensional vectors that form a radiance map describing the view dependent radiance at position $\boldsymbol{p}$, and $\boldsymbol{\beta}$ is a $D-$dimensional vector for the $D$ components of the deep radiance map. Note that $D$ is a hyperparameter here and is set as 8 for most of the scenes in FastNeRF and this work as well. The color at a given 3D position is then computed by taking the inner product of the weights and deep radiance map:
\begin{equation}
\boldsymbol{c} = (r,g,b) = \sum_{i=1}^D\beta_i(u_i,v_i,w_i) = \boldsymbol{\beta}^T \cdot (\boldsymbol{u}, \boldsymbol{v}, \boldsymbol{w}).
    \label{eq:rgb_from_uvw}
\end{equation}

After computing the color $\boldsymbol{c}$ at a given 3D position, the final pixel color can be computed in a similar manner as NeRF by using \cref{eq:vol}. 

\subsection{SqueezeNeRF}
\label{sec:squeezenerf}
\textbf{Default architecture}: In our proposed SqueezeNeRF, we further split $F_{pos}$ into three networks as:

$F_{\text{pos}}^{j}:j \rightarrow \{(\sigma^{j}_0, W_\sigma^{j}), (\boldsymbol{w}_{\boldsymbol{u}}^{j}, \boldsymbol{u}^{j}_0, \boldsymbol{w}_{\boldsymbol{v}}^{j},  \boldsymbol{v}^{j}_0, \boldsymbol{w}_{\boldsymbol{w}}^{j}, \boldsymbol{w}^{j}_0)\}$, for $\forall \ \ j \in \{ (x,y),(y,z),(z,x) \}$.

We then compute $\sigma^j, \boldsymbol{u}^j, \boldsymbol{v}^j$ and $\boldsymbol{w}^j$ ($*$ denotes point-wise multiplication) as:
\begin{equation}
\sigma^{j}=\sigma^{j}_0*w_\sigma^{j}, \boldsymbol{k}^{j}~=\boldsymbol{w}_{\boldsymbol{k}}^{j}*\boldsymbol{k}^{j}_0 \ \ \forall \ \ \boldsymbol{k} \in \{ \boldsymbol{u}, \boldsymbol{v}, \boldsymbol{w}\}.
\label{eq:suvw_computation}
\end{equation}   
Essentially for each $\sigma^j$ and $\boldsymbol{k}^j$, we also compute its corresponding weight. In our experiments, we found that explicitly predicting the weight of the quantities and then, using them for subsequent computation \cref{eq:suvw_fusion} led to better results. We also present the non-weighted network performance as an ablation study in \cref{table:ablation}.

We then combine the output of the three networks:
\begin{equation}
\begin{aligned}
    \sigma~=~f_\sigma(\{ \sigma^j \}), \boldsymbol{k}~=~f_{\boldsymbol{k}}( \{ \boldsymbol{k}^{j} \}) \ \ \forall \ \ \boldsymbol{k} \in \{ \boldsymbol{u}, \boldsymbol{v}, \boldsymbol{w}\}, \\ \forall \ \ j \in \{ (x,y),(y,z),(z,x) \}.
\end{aligned}
\label{eq:suvw_fusion}
\end{equation}

There can be multiple ways in which we represent the network outputs. In the default SqueezeNeRF configuration, for simplicity, we model each $\sigma^j$ as a scalar value and each $\boldsymbol{k}$ as a $D$ dimensional vector ($w_\sigma^{j} \in \mathcal{R}, \boldsymbol{w}_{\boldsymbol{k}}^j \in \mathcal{R}^{D}$). We use a straightforward combination scheme to model $f_\sigma, f_{\boldsymbol{k}}$ as represented in \cref{fig:arc}. We simply multiply all $\sigma^j$ to get our final $\sigma$ value. We add each $\boldsymbol{k}^{j}$ to get $\boldsymbol{k}$.

\textbf{Alternate architecture}: We also evaluate our factorization scheme with another type of architecture. Referred to as "SqueezeNeRF (alt)" in our results, $F_{\text{pos}}$ is split as 

$F_{\text{pos}^j}:j \rightarrow \{\pmb{\sigma}^{j}, \boldsymbol{u}^{j},  \boldsymbol{v}^{j}, \boldsymbol{w}^{j}\}$. 

Here each $\pmb{\sigma}^j \in \mathcal{R}^{D_\sigma}$ and $\boldsymbol{k} \in \mathcal{R}^{D_{\boldsymbol{k}}}$. In our experiments, $D_\sigma$ is set as 12 and $D_{\boldsymbol{k}}$ is set as 8. Finally $\sigma$ and $(\boldsymbol{u},\boldsymbol{v},\boldsymbol{w})$ is computed as:
\begin{equation}
\begin{aligned}
    \sigma = W_\sigma [\text{concat}(\{\pmb{\sigma}^j\})] + b_\sigma, \\
    \boldsymbol{k} = W_{\boldsymbol{k}} [\text{concat}(\{\boldsymbol{k}^j\})] + b_{\boldsymbol{k}},
\end{aligned}
\label{eq:suvw_fusion_alt}
\end{equation}
where $W_\sigma \in \mathcal{R}^{1\times3D_\sigma} , b_\sigma \in \mathcal{R}$, $W_{\boldsymbol{k}} \in \mathcal{R}^{D\times3D_{\boldsymbol{k}}}, b_{\boldsymbol{k}} \in \mathcal{R}^{D}$ are learnable parameters, that are also saved when the neural network is cached and concat refers to concatenation operation.

Please note that while other factorization schemes such as splitting $(x,y,z)$ into $x, y, z$ instead of $(x,y)$, $(y,z)$ and $(z,x)$ is also possible but, we found that the former factorization scheme lead to considerably inferior results. Experimentation with more sophisticated combination functions in \cref{eq:suvw_fusion} is left as part of future work. 

$F_{\text{dir}}$ remains same as that in FastNeRF.  After computing $\sigma, (\boldsymbol{u}, \boldsymbol{v}, \boldsymbol{w})$, we follow \cref{eq:rgb_from_uvw} to compute the color at a 3D point sample and then use \cref{eq:vol} to compute the final pixel color. After computing the final pixel color, SqueezeNeRF is trained in a similar manner as NeRF.
\subsection{Caching}
\label{sec:caching}
The motivation of our proposed architecture is that the cache-size of the mapping learnt by all the $F_{\text{pos}}^{j}$ and $F_{\text{dir}}$ is considerably small. Following similar caching strategy as described in FastNeRF, we define a bounding box that covers the entire scene captured by the NeRF. We sample $n_p$ values along each 3D dimension and $n_d$ number of points for each of the ray direction coordinate $\theta$ and $\phi$. Hence for each $F_{\text{pos}}^{j}$, we store the output of $n_p^2$ values, that is, we store $\{~\sigma^{j},~(\boldsymbol{u}^{j},~\boldsymbol{v}^{j},~\boldsymbol{w}^{j})~\}$ for every ordered pair of $j$. For default architecture, $\sigma^{j}$ is a scalar value and $(\boldsymbol{u}^{j},~\boldsymbol{v}^{j},~\boldsymbol{w}^{j})$ are $D$ dimensional vectors, we store $3n_p^2(3D+1)$ values for $n_p^3$ combination of $(x,y,z)$. Similarly for SqueezeNeRF (alt), we store $3n_p^2(3D+D_\sigma)$ values for $n_p^3$ combination of $(x,y,z)$. For $F_{\text{dir}}$, we store the output for $D$ dimensional vector for the $n_d^2$ combination of $(\theta, \phi)$. So, we store $n_d^2(D)$ output values of $F_{\text{dir}}$. 

In our experiments, we observe that $n_d=64$ for synthetic 360$^\circ$~\cite{nerf} and $n_d=32$ for LLFF dataset~\cite{llff} is sufficient for good results. We further test the cache based inference model for $n_p=512$ (referred to as 512 cache) and $n_p=1024$ (referred to as 1K cache). Based on our experimental observations, we store each $\sigma^{j}$ using 32 bit floating point precision whereas each of $\boldsymbol{u}^{j}, \boldsymbol{v}^{j}, \boldsymbol{w}^{j}$ and $\boldsymbol{\beta}$ are stored using 16 bit floating point precision. In total, our 1K cache takes up 155MB of memory whereas our 512 cache occupies 39MB. A similarly dense 1K cache for FastNeRF would take approximately 54GB, and the corresponding 512 cache would occupy approximately 6.7GB. Also, the corresponding 1K cache and 512 cache size for NeRF would be approximately 35TB and 4.4TB respectively. Alternatively, SqueezeNeRF (alt) is evaluated with $n_p=512,n_d=256$ and all the values are stored using 16 bit floating point precision. So, the 512 cache for SqueezeNeRF (alt) occupies 58MB. Readers are referred to the Supplementary of \cite{fastnerf} for the formulas used to calculate cache sizes for FastNeRF and NeRF. The formula was similarly adapted for cache-size calculation of SqueezeNeRF.  

Also note that as the dense cache corresponding to FastNeRF, NeRF-SH \cite{plenoctrees} and SNeRG \cite{snerg} are quite huge, they save a sparse cache of the neural network mapping based on the scene geometry. While this works fine in case of the scenes used for comparison in this and their work, we believe that this will be less effective for highly dense and feature-rich scenes. On the other hand, SqueezeNeRF enables storage of affordable dense cache. 

\subsection{Implementation}
\label{sec:implementation}

 The SqueezeNeRF training script is based on that of NeRF\cite{nerf}. Apart from the different architecture, the training process is identical to NeRF. 
 
 The position dependent MLPs are modelled using 6 layers with 256 hidden units each whereas the view dependent MLP is modelled with 4 layers and 128 hidden units. As our network has more number of parameters than NeRF, it's training and inference without caching the trained network is slower than that of NeRF. Other training features such as hierarchical sampling and positional encoding and training and testing hyper-parameters are same as that in NeRF.
 
 At inference, similar to NeRF, our method takes a test view as input and predicts the color at each pixel location as described in \cref{sec:squeezenerf}. We sample 256 points along each projected ray. The 3D location and viewing direction of each point is then used to fetch $\sigma^j, \boldsymbol{u}^{j}, \boldsymbol{v}^{j}, \boldsymbol{w}^{j}, \boldsymbol{\beta}$ from the respective cache files, which are then used to compute the color and density using \cref{eq:suvw_fusion} and \cref{eq:rgb_from_uvw}. Finally, the final pixel color is computed using \cref{eq:vol}.
 
 While we use custom CUDA kernels for speeding up inference, our implementation relies on simply querying our dense cache and computing the pixel color via ray marching. We use nearest neighbour interpolation for cache look-up of the outputs of $F_{pos}^{j}$ and bilinear interpolation for sampling from the cache of $F_{dir}$
 
 Unlike FastNeRF, we did not use bounded volume hierarchy based ray tracing or other performance enhancements. Therefore, the performance of our approach can be further improved with similar techniques. 

\section{Experiments, Result and Discussion}
\label{sec:result}

We evaluate our method along with the current approaches on novel view synthesis in terms of quality of rendered images, speed at which the images are generated and memory requirement for any additional data structure such as a cache for neural network output or weights of trained neural network. We use two datasets for this task - NeRF synthetic 360$^\circ$ \cite{nerf} and forward facing Local Light Field Fusion (LLFF) dataset \cite{llff}. 


\begin{table*}[!t]
  \centering
  \begin{tabular}{|l|c|c|c|c|c|c|}
    \hline
    \multirow{2}{*}{Method} & \multicolumn{3}{c|}{Image Generation Quality} & \multirow{2}{*}{Speed [FPS] $\uparrow$} & \multirow{2}{*}{Memory [MB] $\downarrow$}\\
    \cline{2-4}
    & PSNR [dB] $\uparrow$ & SSIM $\uparrow$ & LPIPS $\downarrow$ & & \\
    \hline 
    NeRF \cite{nerf} & 30.23 & 0.946 & 0.050 & 0.04 & \textbf{5}  \\
    FastNeRF (no-cache) \cite{fastnerf} & 29.16 & 0.936 & 0.053 & 0.03 & 28 \\
    NeRF-SH (no-cache) \cite{plenoctrees} & \underline{31.57} & \underline{0.953} & 0.047 & 14 & 12 \\
    JAXNeRF+ Deferred \cite{snerg} & 30.55 & 0.952 & 0.049 & 0.01 & 18 \\
    KiloNeRF \cite{kilonerf} & 31.00 & 0.920 & 0.060 & \textbf{50}  & ~160 \\ 
    AutoInt (8 sections) \cite{autoint} & 25.55 & 0.911 & 0.170 & 0.6  & \textbf{5}  \\
    DIVeR32 (RT) \cite{diver}   & \textbf{32.12} & \textbf{0.958} & \textbf{0.033} & \underline{47} &  68 \\
    SqueezeNeRF (no-cache) & 29.59 & 0.931 & \underline{0.038} & 0.02 & \underline{11} \\
    \hline
    FastNeRF (1K Cache) & 29.97 & 0.941 & 0.053 & 172 & ~16200 \\
    FastNeRF (512 Cache) & N/A & N/A & N/A & \textbf{238} & ~2700 \\
    NeRF-SH (PlenOctree) & \textbf{31.71} & \textbf{0.958} & 0.053 & \underline{168}  & 1900  \\
    SNeRG (PNG) \cite{snerg} & \underline{30.38} & \underline{0.950} & \underline{0.050} & 84  & \underline{87}  \\
    SqueezeNeRF (1K Cache) & 29.61 & 0.921 & \textbf{0.046} & 165  & 155 \\
    SqueezeNeRF (512 Cache) & 28.12 & 0.917 & 0.087 & \underline{168} &  \textbf{39} \\
    \hline
  \end{tabular}
  \caption{\textbf{Results: Comparison of Image Quality, inference speed and required memory for storing cache/network weights for the novel view synthesis of 800x800 image in synthetic $360^\circ$ dataset.}}
  \label{table:recon_table_synthetic}
\end{table*}

\begin{table*}[!t]
  \centering
  \begin{tabular}{|l|c|c|c|c|c|}
    \hline
    \multirow{2}{*}{Method} & \multicolumn{3}{c|}{Image Generation Quality} & \multirow{2}{*}{Speed [FPS] $\uparrow$} & \multirow{2}{*}{Memory [MB] $\downarrow$}\\
    \cline{2-4}
    & PSNR [dB] $\uparrow$ & SSIM $\uparrow$ & LPIPS $\downarrow$ & & \\
    \hline 
    NeRF \cite{nerf} & \underline{26.50} & \underline{0.855} & \underline{0.07} & \underline{0.06} & \textbf{5} \\
    FastNeRF (no-cache) \cite{fastnerf} & \textbf{27.96} & 0.888 & \textbf{0.063} & 0.04 & 28 \\
    JAXNeRF+ Deferred \cite{snerg} & 24.32 & 0.808 & 0.086 & 0.00 & 18 \\
    AutoInt (8 sections) \cite{autoint} & 24.14 & 0.820 & 0.176 & 0.6 & \textbf{5}  \\
    NeX~\cite{nex} & 27.26 & \textbf{0.904} & 0.178 & \textbf{449} & 89 \\
    SqueezeNeRF (no-cache) & 24.32 & 0.808 & 0.085 & 0.03 & \underline{11} \\
    SqueezeNeRF (alt, no-cache) & 26.08 & 0.859 & 0.081 & 0.02 & \underline{11} \\
    \hline
    FastNeRF (768 Cache) & \textbf{26.04} & \textbf{0.856} & 0.085 & \textbf{714} & ~6100 \\
    SNeRG (PNG) \cite{snerg} & \underline{25.63} & \underline{0.818} & 0.183 & 27  & 373  \\
    SqueezeNeRF (1K Cache) & 22.50 & 0.762 & \underline{0.122} & 480  & 155 \\
    SqueezeNeRF (512 Cache) & 21.62 & 0.729 & 0.189 & \underline{484} & \textbf{39} \\
    SqueezeNeRF (alt, 512 Cache) & 23.87  & 0.817 & \textbf{0.071} & 190  & \underline{58}\\
    \hline
  \end{tabular}
  \caption{\textbf{Results: Comparison of Image Quality, inference speed and required memory for storing cache/network weights for the novel view synthesis of 504x378 image in LLFF dataset.}}
  \label{table:recon_table_llff}
\end{table*}

\begin{figure*}[!t]
    \centering
    \begin{subfigure}{\linewidth}
        \centering
        \includegraphics[width=\linewidth]{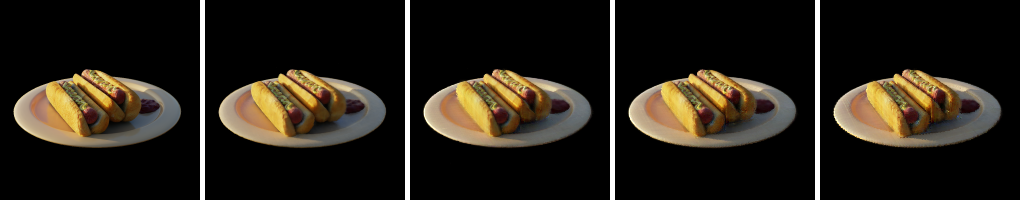}
        \caption{Blender scene: Hotdog}
        \label{fig:qualitative-hotdog}
    \end{subfigure}
    \begin{subfigure}{\linewidth}
        \centering
        \includegraphics[width=\linewidth]{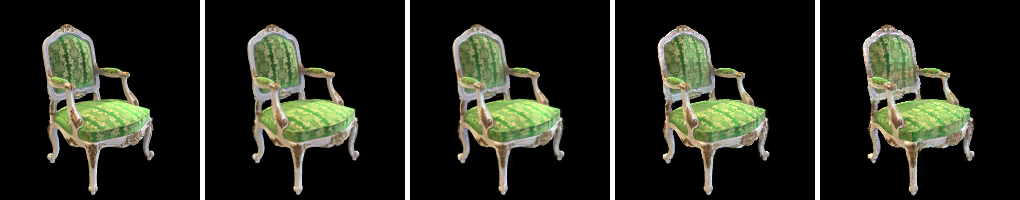}
        \caption{Blender scene: Chair}
        \label{fig:qualitative-chair}
    \end{subfigure}
    \begin{subfigure}{\linewidth}
        \centering
        \includegraphics[width=\linewidth]{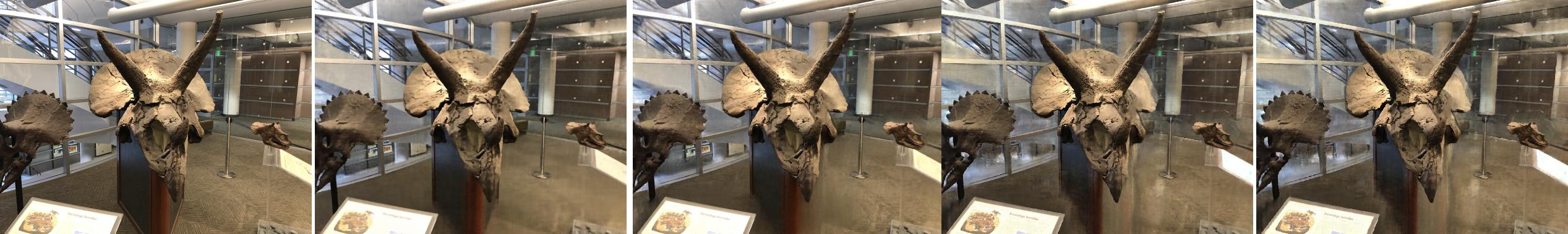}
        \caption{LLFF scene: Horns}
        \label{fig:qualitative-horns}
    \end{subfigure}
    \begin{subfigure}{\linewidth}
        \centering
        \includegraphics[width=\linewidth]{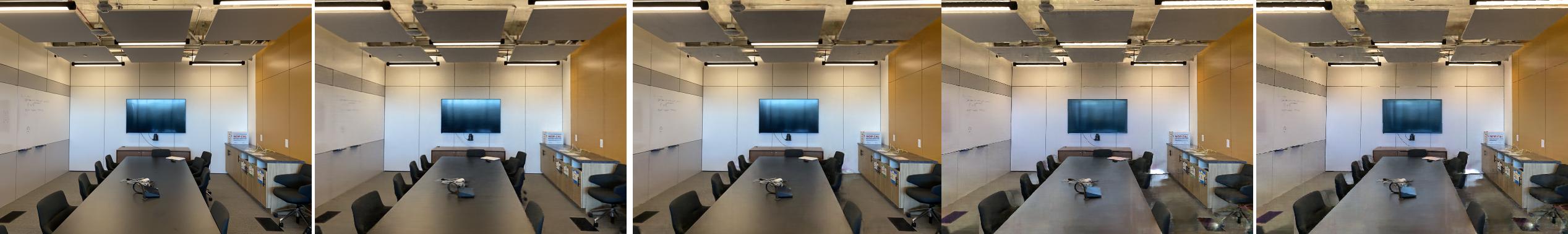}
        \caption{LLFF scene: Room}
        \label{fig:qualitative-room}
    \end{subfigure}    
    
  \caption{\textbf{Results: Images generated from a novel view in the test set.} From left to right: Ground truth image from the test set, NeRF generation, SqueezeNeRF (no-cache) generation, SqueezeNeRF (1K-cache) generation, SqueezeNeRF (512-cache) generation}
  \label{fig:qualitative}
\end{figure*}

\textbf{Image quality}: To evaluate the quality of rendered images, we quantitatively measure the performance by comparing the model output to its corresponding ground truth image in the test set using 3 metrics - Peak Signal to Noise Ratio (PSNR), Structured Similarity (SSIM) \cite{ssim} and LPIPS \cite{lpips}. In  the top half of \cref{table:recon_table_synthetic} and \cref{table:recon_table_llff}, we compare the quality of images rendered by our model (without caching) to NeRF \cite{nerf}, the corresponding network outputs of the different NeRF variants that also rely on caching for speeding up inference \cite{fastnerf, plenoctrees, snerg} and faster inference based NeRF models that do not rely on caching. In the bottom half, we compare the cache based inference of the different NeRF variants to our model's cache based inference. The reported PSNR for NeRF synthetic 360$^\circ$ and LLFF are averaged across all the eight types of scenes present in the two types of datasets. 
KiloNeRF \cite{kilonerf}, DIVeR32 \cite{diver} and NeRF-SH \cite{plenoctrees} are not designed for frontal facing and unbounded scenes so, their performance is not reported in ~\cref{table:recon_table_llff}. Likewise, NeX~\cite{nex} is not designed for 360$^\circ$ scenes so its performance is not reported in \cref{table:recon_table_synthetic}. The reported PSNR in these tables are taken from the respective papers. From these tables, we can see that the quality of images generated by our proposed network architecture is competitive with NeRF and other models. Cache based inference does lead to a degradation in the quality of the images, but also provides significantly high rendering speed. For the NeRF synthetic 360$^\circ$ dataset, 
the drop in quality due to cache based inference is still small (<7\%) and the rendered images in \cref{fig:qualitative} (a-b) also verify that that. For LLFF dataset, the drop in image quality is relatively higher but still, the rendered images in \cref{fig:qualitative} (d-e) still look reasonable. The significant drop in the quality of images generated by SqueezeNeRF for LLFF dataset served as our motivation to propose SqueezeNeRF (alt). From the tables, we can see that the quality of images generated by SqueezeNeRF (alt) is superior to SqueezeNeRF but the performance improvement comes at the cost of slower inference speed. With that said, with caching, we can still render at 190 FPS with SqueezeNeRF (alt).

\textbf{Rendering speed comparison}: We compare the inference speed of our method with the baselines in terms of number of image frames generated per second. In \cref{table:recon_table_synthetic} and \cref{table:recon_table_llff}, the inference speed for NeRF~\cite{nerf}, NeRF-SH (no-cache and plenoctree)~\cite{plenoctrees}, AutoInt~\cite{autoint} has been meaured on a single Nvidia V100 GPU. FastNeRF has been measured on Nvidia RTX 3090, while JAXNeRF+ Deferred and SNeRG have been measured on Nvidia RTX 2080 and KiloNeRF~\cite{kilonerf} and DiVeR32~\cite{diver} have been measured on Nvidia GTX 1080 Ti. For the purposes of comparison in \cref{table:recon_table_synthetic}, \cref{table:recon_table_llff} and \cref{table:cache_generation}, we report our performance on a single Nvidia V100 GPU. We also report our inference speed on Nvidia A100 GPU in \cref{table:v100_a100}.

In \cref{table:recon_table_synthetic} and \cref{table:recon_table_llff}, we can see that our method enables real-time inference of NeRF and is able to render at over 150 frames per second. The rendering speed of our method is competitive with the state of the art methods \cite{fastnerf, plenoctrees}, while our memory requirement is significantly less than these methods. Moreover, in \cref{table:cache_generation}, we also compare \cite{fastnerf, plenoctrees} with our method in terms the time taken to generate taken to generate the cache for neural network. Our smaller cache-size allows us to generate the cache in less than 9s, in contrast to the considerably longer time required by other methods, which is another advantage of our method. So even if the inference speed of our 512 cache model is less than the corresponding FastNeRF model, it will take FastNeRF over 16 minutes to compensate for the longer cache generation time in comparison to our approach.

\textbf{Memory comparison and speed-memory trade-off}: In \cref{table:recon_table_synthetic} and \cref{table:recon_table_llff}, we also compare the memory requirement of our method along with our baselines. For SNeRG \cite{snerg}, NeRF-SH (PlenOctree) \cite{plenoctrees}, FastNeRF~\cite{fastnerf} cache based inference and our method, memory here refers to the network cache-size. For other methods, memory here refers to the memory occupied by the weights of the trained neural network. From the two tables, we can see that non-caching based methods have a considerably less memory overhead but have a slower inference speed compared to the caching based methods. So there is a clear trade-off  between the memory efficiency of a model and it's inference speed. From \cref{table:recon_table_synthetic}, \cref{table:recon_table_llff} and \cref{fig:mem_speed_graph}, we can see that our model can generate images at a high speed, which is competitive with caching based methods, with sufficiently low memory requirement, which is closer to the non-caching based methods.  

\textbf{Other experiments}: In \cref{table:ablation}, we also present results of our model with different configurations. We compare the quality of image generation of our final model configuration (referred to as SqueezeNeRF (no-cache) in \cref{table:ablation}) in terms of PSNR to the different configurations of hidden dimension in the MLPs of $F_{\text{pos}}^{xy}$, $F_{\text{pos}}^{yz}$ and $F_{\text{pos}}^{zx}$ and different settings of $D$, dimensionality of $\boldsymbol{u}, \boldsymbol{v}, \boldsymbol{w}$ and $\boldsymbol{\beta}$. In \cref{table:ablation}, we also show the effect of number of bins per dimension on the quality of images generated via cache-based inference. 

\begin{table}[!t]
  \centering
  \begin{tabular}{|l|c|c|c|}
    \hline
    \multirow{2}{*}{Method} & Cache-size & \multicolumn{2}{c|}{Speed [FPS] $\uparrow$} \\
    \cline{3-4}
    & [MB] $\downarrow$ &  V100 & A100 \\
    \hline 
    SqueezeNeRF (1K cache) & 155 & 165 & 200   \\
    SqueezeNeRF (512 cache) & 39 & 168  & 202  \\
    \hline
  \end{tabular}
  \caption{\textbf{Results: cache-size and inference speed.} SqueezeNeRF cache-size and inference speed of the 800x800 images from synthetic 360$^\circ$ scene. The inference speed has been reported on two different GPUs - Nvidia V100 and Nvidia A100. }
  \label{table:v100_a100}
\end{table}

\begin{table}[!t]
  \centering
  \begin{tabular}{|l|c|}
    \hline
    Method & Time [sec] $\downarrow$ \\
    \hline 
    FastNeRF (1K cache)~\cite{fastnerf} & 420   \\
    NeRF-SH (PlenOctree)~\cite{plenoctrees}  &   1539 \\
    SqueezeNeRF (1K cache) &   9 \\
    \hline
  \end{tabular}
  \caption{\textbf{Results: Cache generation time on a single Nvidia V100 GPU.} Comparison of time required to cache the trained neural network into a sparse 3D grid (for FastNeRF) or Octree (for PlenOctree) or dense 2D grid (for SqueezeNeRF).}
  \label{table:cache_generation}
\end{table}

\begin{table}[!h]
  \centering
  \begin{tabular}{|l|c|}
    \hline
    Configuration & PSNR [dB] $\uparrow$ \\
    \hline 
    NeRF~\cite{nerf} & 26.80   \\
    \hline
    SqueezeNeRF (No-cache): 256 hidden units,  & \multirow{2}{*}{26.92} \\
    D=8, weighted combination \cref{eq:suvw_computation} & \\\hline
    128 hidden units, $D=8$, using \cref{eq:suvw_computation} & 25.61 \\
    256 hidden units, $D=6$, using \cref{eq:suvw_computation} & 26.31 \\
    256 hidden units, $D=10$, using \cref{eq:suvw_computation} & 26.67\\
    256 hidden units, $D=8$, without \cref{eq:suvw_computation} & 24.11\\
    \hline
    $n_p=1024, n_d=64$ (cache-size=155MB) & 26.68 \\
    $n_p=1024, n_d=32$ (cache-size=155MB) & 26.46 \\
    $n_p=512, n_d=64$ (cache-size=39MB) & 24.52 \\
    $n_p=512, n_d=32$ (cache-size=39MB) & 24.42 \\
    \hline
  \end{tabular}
  \caption{\textbf{Results: Other experiments and ablation study.} Comparison of Image quality (in PSNR) of the different SqueezeNeRF configurations for the "Chair" scene in synthetic 360$^\circ$ dataset. In the first row, we report NeRF performance as a reference.}
  \label{table:ablation}
\end{table}

\section{Summary and Limitations}
\label{sec:conclusion}

We present SqueezeNeRF, a further factorized variation of FastNeRF \cite{fastnerf} that allows real-time rendering of NeRF~\cite{nerf} in a memory efficient manner. Similar to the state of the art methods, FastNeRF and NeRF-SH \cite{plenoctrees}, our method also relies on storing a cache of the neural network mapping so that during inference we can replace the millions of neural network computations by simple look-up operations. But while these method can also render at over 150 frames per second, their cache-size, even though they store a sparse version of it, is in the order of few GBs which is a major drawback for any embedded systems application. Our proposed model factorizes the NeRF MLPs into a view conditioned network which takes the camera viewing direction $(\theta, \phi)$ as input and three position conditioned networks, which take $(x,y), (y,z)$ and $(z,x)$ as input respectively. This factorization allows us to reduce the memory complexity of our neural network cache from $\mathcal{O} (n^3)$ (for FastNeRF) to $\mathcal{O}(n^2)$. This allows us to store a dense cache which occupies less than 160MB and still enables us to render over 160 frames per second with Nvidia V100 and over 200 frames per second with Nvidia A100.   

Despite being competitive with the state of the art models in rendering speed with a considerably less memory overhead, the quality of images generated via cache based inference of our approach is inferior to NeRF and some of the other baselines. This can be attributed to our relatively simple combination scheme of fusing the intermediate outputs as described by \cref{eq:suvw_fusion}. A more sophisticated fusion of these intermediate output or incorporation of training techniques from some other follow-up works on NeRF such as \cite{dsnerf, donerf, nerfpp} should lead to higher quality of rendered images. Another point to worth noting is that our factorization scheme, while applied to the position dependent network of FastNeRF in this work, also lends itself for future application to NeRF-SH and SNeRG \cite{snerg} in a similar manner.

{\small
\bibliographystyle{ieee_fullname}
\bibliography{egbib}
}

\end{document}